\pdfoutput=1

\documentclass[11pt]{article}
\usepackage[final]{acl}

\usepackage{times}
\usepackage{latexsym}
\usepackage{multirow}
\usepackage{makecell}
\usepackage{booktabs}
\usepackage{amsmath} 

\usepackage[T1]{fontenc}

\usepackage[utf8]{inputenc}

\usepackage{microtype}

\usepackage{inconsolata}

\usepackage{graphicx}
\usepackage{subcaption}
\usepackage{paralist}

\newcommand{\repo}{\url{https://github.com/zainali93/UrduHMFND2024}}

\title{Detection of Human and Machine-Authored Fake News in Urdu}

\author{
  Muhammad Zain Ali$^{1}$, 
  Yuxia Wang$^{2}$, 
  Bernhard Pfahringer$^{1}$, 
  Tony Smith$^{1}$ \\
  $^{1}$University of Waikato, NZ \, $^{2}$MBZUAI, UAE \\
  \texttt{ma1389@students.waikato.ac.nz}, \texttt{\{bernhard, tcs\}@waikato.ac.nz} \\
  \texttt{yuxia.wang@mbzuai.ac.ae}
}

\begin{document}
\maketitle
\begin{abstract}
The rise of social media has amplified the spread of fake news, now further complicated by large language models (LLMs) like ChatGPT, which ease the generation of highly convincing, error-free misinformation, making it increasingly challenging for the public to discern truth from falsehood. Traditional fake news detection methods relying on linguistic cues also becomes less effective. 
Moreover, current detectors primarily focus on binary classification and English texts, often overlooking the distinction between machine-generated true vs. fake news and the detection in low-resource languages. 
To this end, 
we updated detection schema to include machine-generated news 
with focus on the Urdu language. 
We further propose a hierarchical detection strategy to improve the accuracy and robustness. Experiments show its effectiveness across four datasets in various settings.\footnote{Our data and code are available at \repo.}
\end{abstract}

\section{Introduction}
The aim of fake news detection is to identify false or misleading information presented in news~\cite{shu2019studying}. 
%
%
With the rise of social media, users can post anything virtually without restriction, accelerating the spread of misleading information.
A substantial percentage of content shared on social media is found to be fake, making it challenging for the general public to distinguish truth from falsehood. A recent study revealed that 48\% of individuals across 27 countries have been misled by fake news, mistakenly believing a false story to be true before later discovering its fabrication.\footnote{\url{https://redline.digital/fake-news-statistics/}} 
This phenomenon may lead to serious consequences, including influencing public opinion, undermining democratic processes, and exacerbating societal divisions~\cite{tandoc2018defining,lewandowsky2017beyond}. 
Therefore, effective fake news detection is crucial for maintaining a reliable society and ensuring the integrity of information.

While there have been many studies in English fake news detection, research on low-resource languages such as Urdu remains under-explored \cite{previti2020fake}.  
Previous work has always formulated fake news detection as a binary classification task and identified using linguistic features. 
However, the ease of access to LLMs like GPT-4o~\cite{gpt-4o} has enabled propagandists to produce content that mimics a journalistic tone with minimal errors, complicating the task of determining the veracity of a given text~\cite{wang-etal-2024-m4gt, wang-etal-2024-semeval-2024}.
Additionally, LLMs are increasingly being utilized by journalists and media organizations, further blurring the lines between fake and real news. Currently, the publicly available Urdu datasets consist solely of human-written text \cite{amjad2020bend, akhter2021supervised}, 
which limits the development of existing detection methods. 

To fill this gap, we first collected machine-generated news based on four existing datasets that span short news subtitles and long articles, resulting in four four-label datasets comprising \textit{human fake}, \textit{human true}, \textit{machine fake}, and \textit{machien true}. 
We found that baseline detectors using fine-tuned RoBERTa are not robust, tending to mispredict \textit{machine true} and \textit{machine fake} to other classes. 

To address this, we propose a hierarchical method which breaks down the original four-class problem into two subtasks: machine-generated text detection and fake news detection, as illustrated in Figure \ref{fig:proposed_arch}. Experiments demonstrate that the proposed approach outperforms the baseline across both in-domain and cross-domain settings, in terms of accuracy and F1-scores, highlighting its effectiveness and robustness. Our contributions are summarized as follows:
\begin{compactitem}
    \item We collected the first Urdu dataset for machine-generated fake and true news.
    \item We proposed an effective hierarchical approach for four-label fake news detection, which is more accurate and robust than fine-tuned RoBERTa on four labels.
    \item We conducted a detailed analysis investigating (1) reasons of low accuracy in cross-domain settings, 
    and (2) the impact of data augmentation in machine-generated text detection task on enhancing the overall fake news detection.

\end{compactitem}


\section{Related Works}
This section reviews previous research on 
(1) methods for detecting fake news, with an emphasis on both general approaches and those specific to Urdu, and (2) techniques for generating real and fake machine-generated news.

\paragraph{General Approach for Fake News Detection}
Fake news detectors vary from input features and modeling architectures. 
Features involve content features, social features, temporal features, or combinations of these \cite{shu2017fake}. Content features encompass details such as term frequency~\cite{ahmed2017detection}, sentiment~\cite{bhutani2019fake}, and parts of speech~\cite{balwant2019bidirectional}. Social features are primarily used on social media platforms and include information such as friends' circles, pages followed, and reactions to posts~\cite{sahoo2021multiple}. Temporal features consist of time-related aspects that indicate when a given post was released. For example, \citet{previti2020fake} proposed a Twitter-based fake news detector that incorporated time series data along with other features, reporting favorable results. 

Research has explored various model architectures, ranging from simplistic machine learning (ML) algorithms to advanced transformers~\citep{transformer}. For instance, \citet{raza2022fake} introduced an encoder-decoder transformer that leverages content and social data for early detection. Unsupervised methods aim to circumvent the labor-intensive task of labeling and utilize various heuristics in clustering.
\citet{yin2007truth} suggested that a website's credibility is linked to its consistency in providing accurate information. Similarly, \citet{orlov2019using} proposed a heuristic indicating that coordinated propagandists tend to exhibit similar patterns.

\paragraph{Urdu Fake News Detection} 
Research on the Urdu language is under-explored. 
Existing studies often exhibit a lack of diversity in the features and the model architectures.
\citet{kausar2020prosoul} employed n-grams and BERT embeddings as features, and logistic regression and CNNs as models for training the classifiers. However, translation versions of datasets do not necessarily reflect the actual news lexicon of the target language in real-world scenarios. 
Similarly, \citet{amjad2020data} compared models trained on organically-labeled Urdu fake news data with those trained on English fake news data translated into Urdu. It showed that models trained on organically-labeled Urdu data outperformed those trained on translations.

For Urdu datasets, 
\citet{kausar2020prosoul} translated an English dataset Qprop \cite{barron2019proppy}, into Urdu using Google Translate. 
\citet{akhter2021supervised} created an Urdu fake news dataset by semi-automatic translation of an existing English fake news dataset, and used ensemble approaches and content features for model training. 
In addition, there are three commonly-used fake news datasets that are specifically curated in the Urdu language: \textit{bend-the-truth}~\citep{amjad2020bend}, \textit{ax-to-grind}~\cite{harris2023ax}, and \textit{UFN2023}~\citep{farooq2023fake}. This work also uses these datasets for experiments, and we detailed them in Section \ref{sec:datasets}.

\paragraph{Machine Generated Text} 
What prompts were used to generate paraphrased text via LLMs? 
\citet{zellers2019defending} trained a model GROVER, which could generate and identify the fabricated articles. 
\citet{huang2022faking} used BART for mask-infilling to replace salient sentences in articles with plausible but non-entailed text, ensuring disinformation through self-critical sequence training with an NLI component. Similarly, \citet{mosallanezhad2021generating} proposed a deep reinforcement learning-based method for topic-preserving synthetic news generation, controlling the output of large pre-trained language models. 
All of these studies focused on generating fake news. However, LLMs are now utilized by news organizations and journalists, requiring a new schema of generating machine true news.
\citet{su2023fake} presented \textit{Structured Mimicry Prompting} approach for generating both machine fake and true news using \textit{GPT-4o}, in which LLM understands the title and article body, and generate a similar text. 

\begin{table*}[t!]
\centering
\resizebox{\textwidth}{!}{%
\begin{tabular}{lrrrrr | rrrr | rr}
\toprule
\textbf{Dataset} & \#\textbf{Examples} & \#\textbf{HF} & \#\textbf{HT} & \#\textbf{MF} & \#\textbf{MT}  & \textbf{$\hat{T}$(HF)} & \textbf{$\hat{T}$(HT)} & \textbf{$\hat{T}$(MF)} & \textbf{$\hat{T}$(MT)} & \textbf{Content} & \textbf{Category} \\ \midrule
Dataset1         & 10083  & 5053 & 5030     & 5053 & 5030   & 58.7 & 19.2 & 61.2 & 20.1 &  headlines & Short             \\
Dataset2         & 4097   & 2455 & 1642     & 2455 & 1642   & 105.6 & 34.3 & 110.2 & 33.4 &  headlines & Short             \\
Dataset3         & 2000   & 968  & 1032     & 968  & 1032   & 645.0 & 516.1 & 602.2 & 499.4 & articles  & Long              \\
Dataset4         & 1300   & 550  & 750      & 550  & 750    & 134.1 & 198.0 & 101.3 & 211.6 & articles  & Long              \\ 
\bottomrule
\end{tabular}%
}
\caption{Statistical information of four datasets. Examples are organic samples. \#= the number of news, $\hat{T}$=average tokens. \textbf{HF}: Human Fake, \textbf{HT}: Human True, \textbf{MF}: Machine Fake, \textbf{MT}: Machine True.}
\label{tab:datasets} 
\end{table*}
\section{Dataset Collection}

\subsection{Datasets}
\label{sec:datasets}
For training the models, four publicly available Urdu fake news datasets have been used, with the new creation of data for two classes: \textit{machine true} and \textit{machine fake}. This section provides a description and analysis of these four datasets.
\paragraph{Dataset 1: \textit{Ax-to-Grind Urdu}}
The latest Urdu fake news dataset that was presented earlier this year is \textit{Ax-to-Grind Urdu}. It has 10083 samples related to fifteen different domains with approximately equal distribution of fake and true classes. \citet{harris2023ax} maintained the originality of the corpus by keeping only the original news articles. For real news, the data was collected from authentic news websites such as BBC Urdu, Jang, Dawn news etc. Fake news data was collected from two of the arguably most controversial news websites: Vishwas News and Sachee Khabar. Additionally, some fake news was collected through crowd sourcing. Professional journalists were hired to fact-check each individual news sample and label it accordingly.
\paragraph{Dataset 2: \textit{UFN2023}}
This dataset was constructed using a hybrid approach that involved authentic news websites for real news and supervised translation of samples from an English dataset in the fake category into Urdu for the fake news. Additionally, some obvious fake news articles from \textit{Vishwas News} were also included. The dataset contains 4,097 samples across nine different domains, such as health, sports, technology, and showbiz. Of these, 1,642 samples belong to the real news category, while 2,455 belong to the fake news category. 
\paragraph{Dataset 3: \textit{UFN Augmented Corpus}}
\textit{UFN Augmented Corpus} is another publicly available Urdu Fake News dataset. \citet{akhter2021supervised} randomly selected two thousand news articles from an English fake news dataset and performed automatic translation using Google Translate. The name of the English dataset has not been revealed in their work. Out of two thousand translated articles, 968 news articles belong to the fake class and 1032 news article belong to the true class.
\paragraph{Dataset 4: \textit{Bend the Truth}}
This is arguably one of the first publicly available Urdu fake news datasets presented by \citet{amjad2020bend}. It is a relatively small dataset consisting of only 1300 (750 real, 550 fake) news articles, but the authors used a very interesting approach to keep the dataset organic. They collected true articles from authentic news websites and then they hired journalists 
to rewrite these news articles with a counter-factual narrative while keeping the original journalistic tone.    

\paragraph{Analysis \& Categorization} 
Table \ref{tab:datasets} provides a summary of the four datasets used in this work. To determine the number of tokens, stop words were removed, and the text was tokenized using the word tokenizer from the NLTK library, which, for Urdu, segments words based on spaces. Based on the average lengths in both categories, the datasets are classified as either Short or Long. This classification is determined by the token lengths: Datasets 1 and 2 primarily contain short texts and news headlines, thus categorized as \textit{Short}, whereas Datasets 3 and 4 consist of longer news articles, thus categorized as \textit{Long}.
\begin{figure}[t!]
  \centering
  \includegraphics[width=0.48\textwidth]{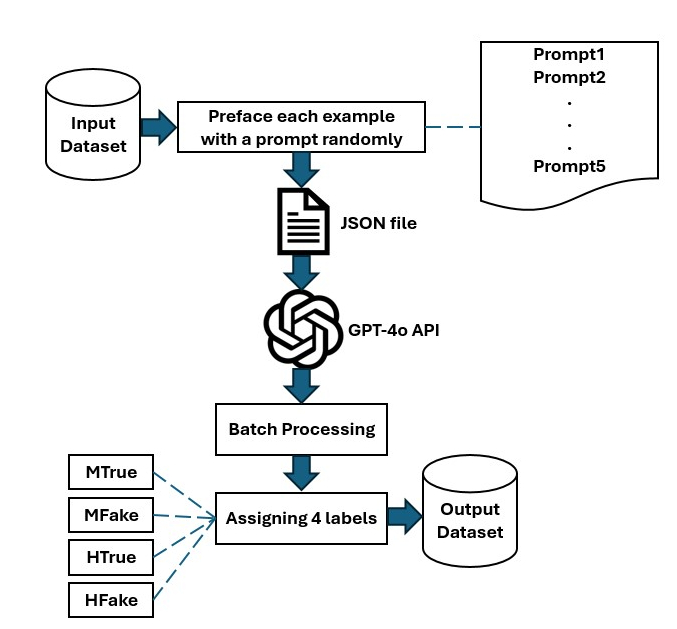}
    \caption{Machine generated News collection process}
  \label{fig:mgt}
\end{figure}

\paragraph{Split of Training, Development and Test Sets}
To obtain a completely unseen test set for evaluation purposes, a 20\% stratified random split was set aside for each of the four datasets. This ensures that even the machine-generated text remains entirely unseen by the trained classifiers. The validation set used for each experiment consisted of 25\% of the training set. In summary, 60\% of the data is used for training, 20\% for validation, and 20\% for testing.

\subsection{Machine-generated News Collection}
GPT-4o was used with five different prompts to generate paraphrased news articles and short messages for both true and fake categories, effectively doubling the number of examples in all four datasets. Figure \ref{fig:mgt} shows the process employed to perform dataset addition and creation of new labels. Each example in the dataset is randomly prefaced with one of the five prompts and saved as an item in a JSON file, including other necessary fields for batch processing, such as \textit{url}, \textit{method}, \textit{body}, \textit{model}, and \textit{max\_tokens}. The resulting JSON file is then sent to the OpenAI client to generate the required data. In the final step, labels are assigned such that the original articles are changed from True and Fake to \textit{human true} and \textit{human fake}. Machine-generated articles receive \textit{machine true} and \textit{machine fake} labels based on the labels of their parent news articles.

\paragraph{Generation Prompts}
Table \ref{tab:prompts} shows all the five prompts used for generating the additional data. The first prompt was created in a careful manner to instruct GPT-4o to rephrase the provided article by keeping the exact same stance. Other prompts were created by GPT-4o itself by giving the following prompt: \textit{Please help create multiple similar prompts like the following:<original prompt>}.
\begin{table*}[t!]
\centering
\resizebox{\textwidth}{!}{%
\begin{tabular}{lp{0.85\textwidth}c}
\toprule
\textbf{Prompt} & \textbf{Text} & \textbf{Author} \\ 
\midrule
\text{Prompt1} & \textit{I am going to provide you with an Urdu article. Please rewrite that article while keeping the same narrative. Feel free to completely change everything, every single word if you have to. In fact, I would appreciate it if there is very little similarity between the original article and what you write. Just the idea and narrative should essentially be the same. The article follows:} & Human \\
\midrule
\text{Prompt2} & \textit{I will provide you with an Urdu article. Your task is to rewrite this article while maintaining the same core message and narrative. Ensure that the wording and structure are significantly different from the original. Here is the article:} & GPT-4o \\
\midrule
\text{Prompt3} & \textit{Here is an Urdu article that I need you to rewrite. Please keep the underlying story and narrative intact, but rephrase it thoroughly so that it appears entirely new. Aim for minimal similarity to the original text. The article is as follows:} & GPT-4o \\
\midrule
\text{Prompt4} & \textit{Please take the following Urdu article and rewrite it in such a way that the narrative and main idea remain unchanged, but the language and wording are entirely different. Your goal is to create a version with minimal resemblance to the original. Here is the article:} & GPT-4o \\
\midrule
\text{Prompt5} & \textit{Given the following Urdu article, I need you to produce a rewritten version that preserves the same story and narrative. Feel free to alter the wording and sentence structure extensively to ensure the new version is distinct from the original. The article is:} & GPT-4o \\ 
\bottomrule
\end{tabular}%
}
\caption{Different prompts used for rewriting Urdu articles while maintaining the core narrative.} 
\label{tab:prompts}
\end{table*}
\paragraph{Quality Control}
Aforementioned prompting strategy was tested on the first dataset and detailed analysis was performed on the resultant machine-generated news articles in following two ways: (i) a native Urdu speaker author reviewed random machine-generated samples to compare them with original articles and report any discrepancies; and (ii) The final dataset, comprising both human-generated and machine-generated news articles, was tokenized. A comparison was made between the number of tokens in the original and machine-generated versions, applying a 20\% threshold to filter out examples with that percentage difference.

Three problems were identified: (i) Some examples were not paraphrased; instead, GPT-4o responded with prompts like: \textit{Please provide the news article for rephrasing.} (ii) For a few short examples, GPT-4o hallucinated information not present in the original text. (iii) Some paraphrased articles began with a preface from GPT-4o, such as: \textit{Certainly! I can help you with rephrasing.}
To address these problems, all prompts were re-engineered by adding the following line before the last sentence: \textit{Please directly rewrite without opening words like Of course I can help you with rewriting, and note that do not generate or extend extra information that is not included in the given article, DO NOT HALLUCINATE EXTRA INFORMATION.} These newly engineered prompts were applied to the problematic samples of the first dataset and the remaining three datasets. The generated text was re-evaluated using the methods discussed above, and no issues were identified in any of the final four labeled datasets.
  
\section{Methods}
This section presents the baseline methods and our proposed hierarchical architecture.
\subsection{Baselines}
The baselines were established in consideration with the following questions: (i) What performance can simpler, hyperparameter-tuned machine learning models achieve on the datasets? (ii) What performance can be obtained by fine-tuning a publicly available multilingual language model for the downstream multiclass classification task? The following baselines were selected:

\paragraph{Linear Support Vector Machines.} 
Several machine learning models were considered, including SVM, Multinomial Naive Bayes, and Random Forest. The overall pipeline begins with data cleaning, which involves removing punctuation, stop words, and URLs. This is followed by bag-of-words representation and TFIDF feature extraction. Models were trained with various hyperparameters, and the best ones were selected using grid search. Among all the models, the Linear Support Vector Machine yielded the best results, with varying cost parameters for each dataset.
\paragraph{Finetuned Xlm-ROBERTa-base.} 
For the downstream multiclass classification task, xlm-ROBERTa-base was selected due to its multilingual capabilities, including understanding Urdu. The text was tokenized, and the four labels were numerically encoded. The pretrained language model was finetuned using a learning rate of $2\times10^{-5}$, weight decay of 0.01, and 10 epochs. The parameter \textit{load\_best\_model\_at\_end} was set to \texttt{True} to retrieve the best model from all epochs. This finetuned model outputs logits for the test data, which are converted to probabilities using the softmax function. The class label is assigned by selecting the index of the maximum probability.
\subsection{Hierarchical Fake News Detection}
From the results of the baseline models in Table \ref{tab:result_indvidual}, it is evident that the performance of the \textit{machine true} and \textit{machine fake} classes is consistently worse than that of their \textit{human true} and \textit{human fake} counterparts. This indicates that machine-generated news is not effectively detected using multiclass methods, highlighting an opportunity for improvement. Intuitively, in the overall four-class fake news classification problem, two requirements emerge: (i) predicting whether a given sentence is written by a machine or a human (machine-generated text detection) and (ii) determining whether an example is fake or not (fake news detection). The baseline results suggest that these two tasks are likely not being addressed efficiently by four-class classification models, be it machine learning or fine-tuned pretrained models.  
To address this problem, a hierarchical method has been proposed that breaks down the multiclass problem into two sub-tasks: machine-generated text detection and fake news detection. Therefore, the focus is on improving the performance of these individual models, which will ultimately enhance the overall results.
\begin{figure*}[ht]
  \centering
  \includegraphics[width=0.8\textwidth]{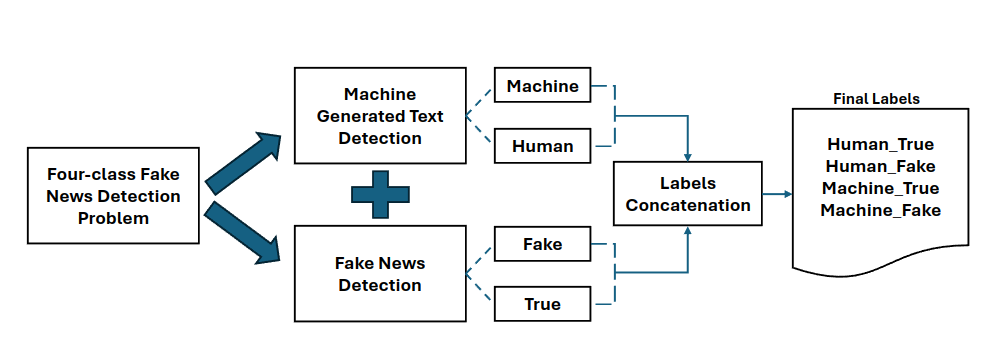}
    \caption{Proposed Hierarchical Fake News Detection Architecture}
  \label{fig:proposed_arch}
\end{figure*}
Figure \ref{fig:proposed_arch} presents an overview of the method employed in this work. To train the model, the training labels are adapted to meet the models' requirements: for machine-generated text detection, only the `Human' and `Machine' portions of the labels are retained, while for the fake news detection model, `Fake' and `True' portions are kept, discarding the rest in both cases. As the base model for finetuning on the downstream machine-generated text detection and fake news detection tasks, xlm-ROBERTa-base is used to ensure a fair comparison with the baseline. The hyperparameters are consistent with those described in the baseline models section. These models are trained and optimized on the validation data accordingly. At inference, both models predict their respective labels for the test data, which are then concatenated and transformed back into four labels.

\section{Experiments}
In this section, we first discuss the experiments results in various settings,
and analyze some interesting findings at the end. 

\subsection{Four-class vs. Hierarchical Detection}
\begin{figure*}[ht]
    \centering
    \begin{subfigure}[b]{0.45\textwidth}
        \includegraphics[width=\textwidth]{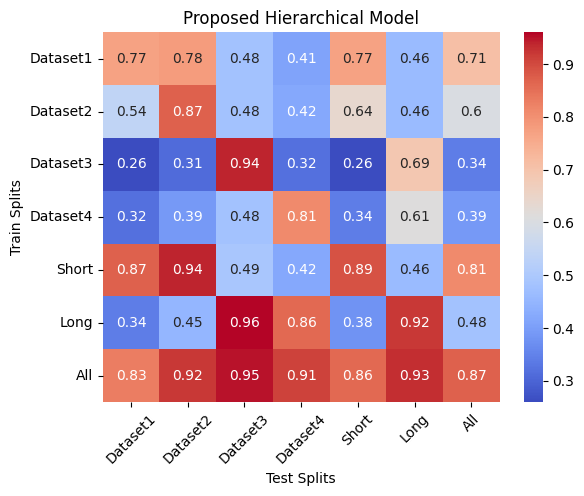}
        \caption{Proposed Model}
        \label{subfig:proposedmodel}
    \end{subfigure}
    \hspace{0.001\textwidth}
    \begin{subfigure}[b]{0.45\textwidth}
        \includegraphics[width=\textwidth]{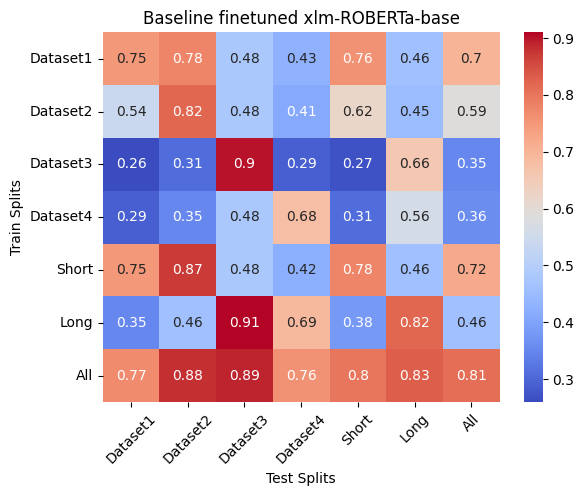}
        \caption{Baseline Model}
        \label{subfig:baseline}
    \end{subfigure}
    
    \caption{Cross-domain evaluation results in terms of \textit{Accuracy}}
    \label{fig:crossdomaineval}
\end{figure*}

\paragraph{Experimental Setup:}
Various experiments were conducted to compare the performance of the proposed hierarchical fake news model against the four-class baseline models. This includes (i) performance on the four datasets used in this work, (ii) a comparison of models trained on the datasets in the \textit{long} category versus those trained on the \textit{short} category, and (iii) the performance of the model trained on all four datasets combined. The test split, set aside at the beginning, was used to measure the performance of all these models (discussed in Section \ref{sec:datasets}). Table \ref{tab:result_indvidual} summarizes the results of these experiments, presenting the overall accuracy and F1 score for the four classes. Additionally, cross-domain evaluation have also been performed which includes how the models trained on one type of dataset performs on the other datasets.
\paragraph{Individual Datasets}
This section analyzes the performance of hierarchical fake news detection models trained on individual datasets (1 to 4) and tested on their respective test splits, comparing them to the baseline models. From Table \ref{tab:result_indvidual}, it is clear that the proposed model consistently outperforms the baselines across all four datasets, in terms of both accuracy and F1 scores for each class. The only exception is dataset 1, where the proposed model ranks a close second for the F1 score of the \textit{human true} class but surpasses in all other F1 scores and overall accuracy. Another key improvement is the reduced gap between the F1 scores of \textit{human fake} and \textit{machine fake}, as well as \textit{human true} and \textit{machine true}. While the baseline models showed a significant difference in F1 scores between \textit{Human} and \textit{Machine} for fake news detection, the proposed hierarchical model has largely bridged this gap, achieving almost identical performance, demonstrating the efficacy of the hierarchical classification approach.
\begin{table}[th!]
\centering
\resizebox{\columnwidth}{!}{%
\begin{tabular}{l | l | ccccc}
\toprule
\textbf{Dataset} & \textbf{Model}                    & \textbf{HF} & \textbf{HT} & \textbf{MF} & \textbf{MT} & \textbf{Acc} \\ 
\midrule
\multirow{3}{*}{Dataset1}               & \textit{LSVM}             & 0.73             & 0.61             & 0.64             & 0.52             & 0.63              \\ 
                                        & \textit{XLM-R-base} & 0.83             & \textbf{0.71}    & 0.77             & 0.69             & 0.75              \\ 
                                        & \textit{Hierarchical}            & \textbf{0.85}    & 0.69             & \textbf{0.8}     & \textbf{0.74}    & \textbf{0.77}     \\ 
                                        \midrule
\multirow{3}{*}{Dataset2}               & \textit{LSVM}             & 0.82             & 0.6              & 0.77             & 0.53             & 0.71              \\ 
                                        & \textit{XLM-R-base} & 0.93             & 0.66             & 0.88             & 0.7              & 0.82              \\ 
                                        & \textit{Hierarchical}            & \textbf{0.93}    & \textbf{0.8}     & \textbf{0.9}     & \textbf{0.77}    & \textbf{0.87}     \\ 
                                        \midrule
\multirow{3}{*}{Dataset3}               & \textit{LSVM}             & 0.89             & 0.87             & 0.86             & 0.85             & 0.87              \\ 
                                        & \textit{XLM-R-base} & 0.91             & 0.91             & 0.88             & 0.89             & 0.9               \\ 
                                        & \textit{Hierarchical}            & \textbf{0.96}    & \textbf{0.95}    & \textbf{0.92}    & \textbf{0.91}    & \textbf{0.94}     \\ 
                                        \midrule
\multirow{3}{*}{Dataset4}               & \textit{LSVM}             & 0.56             & 0.59             & 0.3              & 0.42             & 0.48              \\ 
                                        & \textit{XLM-R-base} & 0.76             & 0.73             & 0.58             & 0.65             & 0.68              \\ 
                                        & \textit{Hierarchical}            & \textbf{0.85}    & \textbf{0.85}    & \textbf{0.74}    & \textbf{0.79}    & \textbf{0.81}     \\ 
\toprule
\multirow{2}{*}{Short}                  & \text{XLM-R-base} & 0.88             & 0.68             & 0.83             & 0.72             & 0.78              \\ 
                                        & \text{Hierarchical}              & \textbf{0.93}    & \textbf{0.85}    & \textbf{0.91}    & \textbf{0.86}    & \textbf{0.89}     \\ 
                                        \midrule
\multirow{2}{*}{Long}                   & \text{XLM-R-base}   & 0.89             & 0.88             & 0.74             & 0.77             & 0.82              \\ 
                                        & \text{Hierarchical}              & \textbf{0.94}    & \textbf{0.94}    & \textbf{0.89}    & \textbf{0.9}     & \textbf{0.92}     \\ 
                                        \midrule
\multirow{2}{*}{All}                    & \text{XLM-R-base}   & 0.89             & 0.77             & 0.83             & 0.74             & 0.81              \\ 
                                        & \text{Hierarchical}              & \textbf{0.91}    & \textbf{0.85}    & \textbf{0.88}    & \textbf{0.83}    & \textbf{0.87}     \\ 

\bottomrule
\end{tabular}
}
\caption{Accuracy (Acc) and F1-score over four labels on four individual datasets (top four rows) and their combinations: Short=1+2, Long=3+4 and All=1+2+3+4. \textbf{HF}: Human Fake, \textbf{HT}: Human True, \textbf{MF}: Machine Fake, \textbf{MT}: Machine True.}
\label{tab:result_indvidual} 
\end{table}
\paragraph{Combinations of Datasets}
This section presents the results for different dataset combinations: \textit{long datasets (dataset 3+4)}, \textit{short datasets (dataset 1+2)}, and \textit{all datasets combined}. Similar training steps and hyperparameters were used, with the training splits of datasets 1 and 2 combined for the \textit{short dataset} model, datasets 3 and 4 combined for the \textit{long dataset} model, and all datasets combined for the \textit{all datasets} model. The baseline LSVM, due to consistently lower performance, was excluded from further experiments.
The bottom three rows of Table \ref{tab:result_indvidual} summarize the performance of models trained on \textit{short}, \textit{long}, and \textit{all} datasets combined. As expected, the proposed model outperformed the baseline across all combined datasets, whether \textit{short}, \textit{long}, or \textit{all}.
Secondly, the proposed models narrowed the gap between F1 scores of \textit{Human} and \textit{Machine} in fake news detection compared to the baseline. In comparing performance, models trained on shorter datasets outperform those trained on longer datasets. This is likely because machine detection is easier on longer datasets, as GPT-4o's rephrasing of longer articles results in higher token variation, allowing the classifier to better distinguish between machine and human text. Similarly, for the all-dataset trained model, the proposed model outperformed the baseline by 6\% in accuracy and achieved higher F1 scores for machine-generated fake news detection.
\paragraph{Cross-domain Evaluation}
Cross-domain evaluation was conducted by testing a model trained on one dataset's training set while testing on test splits of other datasets, resulting in a total of 49 evaluations. In this section, the results are presented only based on accuracy measures. Figures \ref{subfig:proposedmodel} and \ref{subfig:baseline} display the heatmaps of accuracy values for all 49 combinations of the proposed and baseline models, respectively. The y-axis represents the training splits and the x-axis represents the test splits of the datasets. The overall trend of both heatmaps are similar, with higher values along the diagonal and lower values in the non-diagonal entries, except for the last entry of the all-dataset-trained model, which technically cannot be regarded as cross-domain evaluation. This indicates that the models do not generalize well for out-of-domain datasets. However, shorter datasets exhibit relatively better generalization. This trend is less pronounced for longer datasets. Using our proposed method, the model trained on dataset 3 achieves only 32\% accuracy when tested on dataset 4, while the model trained on dataset 4 gains 48\% when tested on dataset 3, showing poor performance.



\subsection{Analysis}
In this section, interesting results from the presented work are analyzed, including the reasons behind the lowest accuracy in cross-domain evaluation when short datasets are used for training and long datasets for testing, and vice versa. Additionally, we examine the results of the experiment aimed at enhancing the model's performance on the first dataset, which exhibited the lowest performance among the individual dataset models.
\paragraph{Low Accuracy in Cross-domain Evaluation using \textit{Short} for Training and \textit{Long} for Testing.} 
    
\begin{figure}[t!]
    \centering
    \includegraphics[scale=0.33]{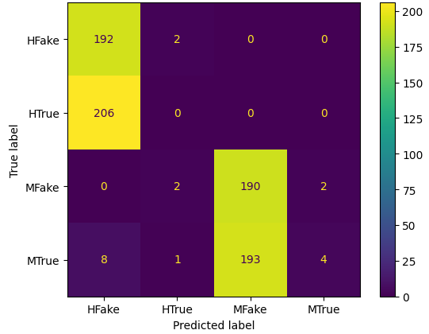}
    \includegraphics[scale=0.33]{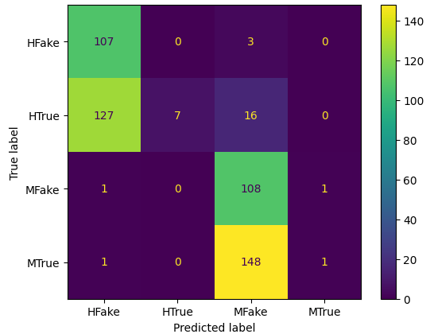}
    \caption{Confusion matrix of testing on long datasets using model trained on dataset1. \textbf{Left:} Test Split Dataset 3 (Long) and\textbf{ Right:} Test Split Dataset 4 (Long)}
    \label{fig:cm10kLong}
\end{figure}

\begin{figure}[t!]
    \centering
    \includegraphics[scale=0.33]{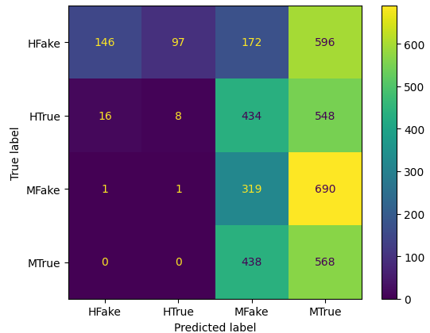}
    \includegraphics[scale=0.304]{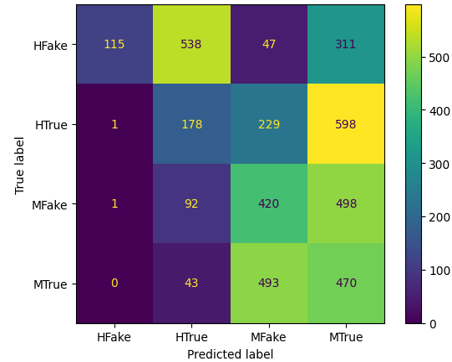}
    \caption{Confusion matrix of testing on dataset 1 using model trained \textbf{Left:} Train Split Dataset 3 (Long) and\textbf{ Right:} Train Split Dataset 4 (Long)}
    \label{fig:cmLong10k}
\end{figure}
    
Figure \ref{fig:cm10kLong} shows the confusion matrices of the model trained on dataset 1, tested on datasets 3 and 4 respectively. Notably, the matrices exhibit almost no correct predictions for the \textit{machine fake} and \textit{machine true} classes. Interestingly, despite overall incorrect predictions, \textit{machine true} is mostly misclassified as \textit{machine fake}, and \textit{human true} as \textit{human fake}, suggesting that the machine-generated text detection component performs well on both long datasets. The reason the fake news detection module fails in this case can be attributed to a simple observation: for short datasets, there is a significant difference in average token count between true and fake classes, with fake articles having more tokens, as shown in Table \ref{tab:datasets}. This may inadvertently cause the model to treat text length as a key feature, resulting in all long articles being classified as fake.

\paragraph{Low Accuracy in Cross-domain Evaluation using \textit{Long} for Training and \textit{Short} for Testing.} 
The models trained on long-category datasets exhibit different behavior when predicting on short datasets. Figure \ref{fig:cmLong10k} shows the confusion matrix of both long datasets-trained models on test split of Dataset 1. Notably, the machine-generated text detection module performs less effectively than in the previous case, with values scattered across the confusion matrix. Secondly, unlike the short datasets, the average tokens for true and fake classes are similar preventing the model from using length as a distinguishing parameter during training. Consequently, the fake news detection module does not classify all short texts as fake. For the model trained on dataset 3, the \textit{human true} class performs poorly, with all samples misclassified as either \textit{machine true} or \textit{machine fake}. In contrast, the model trained on dataset 4 shows a more dispersed confusion matrix with most samples being classified into one of the following three classes: \textit{human true}, \textit{machine true}, or \textit{machine fake}. While precision for \textit{human fake} is high, the recall is low, making it less useful. Overall, this model appears to produce somewhat random predictions, likely due to being trained on the smallest dataset among the four.

\paragraph{Improvement in Dataset 1 by Data Augmenting in MGT Detection.} 
Among the individual datasets, dataset 1 had the lowest performance, with the proposed model achieving an accuracy of only 77\%. Closer inspection revealed that the subpar performance of the machine-generated-text detection module affected the overall results. This may be due to the fact that when GPT-4o rephrases short texts, like those in dataset 1, it makes minimal changes, making it challenging for the model to learn distinguishing features. To test the hypothesis that enhancing machine-generated-text detection would improve overall results, the Urdu subset of a publicly available machine-generated text detection dataset M4~\citep{wang-etal-2024-m4} was augmented, and the model was retrained. This led to a 3\% improvement in the accuracy of the MGT module, which subsequently boosted the overall accuracy of the model trained on dataset 1 by 4\%. This emphasizes the importance of enhancing machine-generated-text detection for the four-label fake news detection, especially for datasets with short texts.

\section{Conclusion and Future Work}
In this work, we introduced a four-class Urdu fake news detection task and presented the first publicly available datasets for this task. We proposed a hierarchical approach that breaks down the four-class problem into machine-generated-text detection and fake news classification. Experiments demonstrate that our approach consistently enhances the accuracy compared to baseline methods and demonstrates robustness across unseen domains. Moreover, the proposed method effectively bridges the gap between F1-score of the \textit{machine true} and \textit{human true} classes, as well as \textit{machine fake} and \textit{human fake} classes, thereby improving the identification of machine-generated fake news. in addition, data augmentation for the machine-generated text (MGT) module improved MGT accuracy and thus enhanced the overall performance for four-class fake news detection. 

For future work, we will explore methods mitigating the classifier from learning length as a feature during training. Additionally, experiments with other multilingual LLMs could further enhance the performance of fake news detection models. Exploring domain adaptation techniques to improve generalization across diverse datasets and integrating explainability methods to understand model decisions are also interesting. 


\section*{Limitations}

We acknowledge certain limitations in this work that can be addressed in future research. First, the relience on publicly available datasets may limit the diversity and richness of the training data, potentially affecting the generalizability of our model. This could lead to suboptimal performance when applied to real-world scenarios where misinformation varies widely in style and content. Secondly, the TFIDF features used for the LSVM classifier may not be the most optimal for fake news detection. Alternative features, such as those derived from the News Landscape (NELA), could enhance performance, but their implementation requires considerable effort, particularly for the Urdu language. Third, the model may inadvertently learn to rely on text length as a distinguishing feature, which could skew predictions, especially with varying lengths of articles. This tendency was observed during the analysis of our results, indicating that further refinement is necessary to mitigate this issue. Finally, the machine-generated text detection (MGT) module primarily addresses a subset of machine-generated content, potentially missing other forms of automated misinformation. Future work could focus on expanding the MGT module to encompass a broader range of machine-generated texts.


\section*{Ethical Statement and Broad Impact}
\paragraph{Ethical Statement}
We recognize that our approach to fake news detection involves the use of machine-generated text, which may inadvertently incorporate biases present in the training data or models. Given the potential for misinformation to influence public opinion and societal well-being, it is crucial to emphasize the importance of human oversight in the evaluation of our system's outputs. We advocate for the involvement of human reviewers, particularly in sensitive contexts, to ensure responsible decision-making and to mitigate the risk of misclassification.
\paragraph{Broader Impact} This work has the potential to significantly enhance the field of fake news detection, particularly for low-resource languages like Urdu. By providing publicly available datasets and a robust hierarchical approach, this research will empower journalists, researchers, and the general public to identify and combat misinformation more effectively. The proposed methodology can be adapted for various applications, including integration into news platforms and social media, thereby facilitating the identification of misleading information and contributing to the overall integrity of public discourse. Ultimately, this work aims to foster a more informed society by improving the tools available for discerning fact from fiction in the rapidly evolving digital landscape.

\bibliography{custom}

\end{document}